%% file: vgan.tex
\documentclass[10pt,twocolumn,letterpaper]{article}

\usepackage{cvpr}
\usepackage{times}
\usepackage{epsfig}
\usepackage{graphicx}
\usepackage{amsmath}
\usepackage{amssymb}
\usepackage{booktabs, makecell}
\usepackage{multirow}
\usepackage{cancel}
\usepackage[normalem]{ulem}
\usepackage[T1]{fontenc}
\usepackage{mathtools}
\newcommand*{\bfrac}[2]{\genfrac{[}{]}{0pt}{}{#1}{#2}}
\usepackage[numbers]{natbib}
\usepackage{color}

\usepackage[final]{animate} 
\usepackage{graphicx}
\usepackage{url}

\usepackage{tabularx, booktabs, ragged2e}
\usepackage{enumitem}
\usepackage{multirow}

\usepackage[percent]{overpic}
\usepackage{xcolor}
\usepackage{pict2e}

\usepackage{caption}
\usepackage{subcaption}

\definecolor{mygreen}{HTML}{008040}
\definecolor{myblue}{HTML}{0040E8}
\definecolor{myred}{HTML}{D20000}

\newif\ifsubmit

\submitfalse

\ifsubmit
\newcommand{\JK}[1]{}
\newcommand{\my}[1]{}
\newcommand{\revmy}[1]{}
{}
\else
\newcommand{\JK}[1]{{\bf \textcolor{red}{Jan: #1}}}
\newcommand{\my}[1]{{\bf \textcolor{magenta}{M.-Y Liu: #1}}}
\newcommand{\revmy}[1]{\textcolor{magenta}{#1}}

\fi

\usepackage[pagebackref=true,breaklinks=true,letterpaper=true,colorlinks,bookmarks=false]{hyperref}

 \cvprfinalcopy

\def\cvprPaperID{1622} 

\newcommand{\ours}{MoCoGAN }

\begin{document}

\title{MoCoGAN: Decomposing Motion and Content for Video Generation}


\author{Sergey Tulyakov,\\
	Snap Research\\
	{\tt\small stulyakov@snap.com}
	\and
	Ming-Yu Liu, \quad Xiaodong Yang, \quad Jan Kautz\\
	NVIDIA\\
	{\tt\small \{mingyul,xiaodongy,jkautz\}@nvidia.com}
}

\maketitle


\input{intro.tex}

\input{rel.tex}
\input{body.tex}

\input{expr_refined.tex}
\input{conc.tex}

{\small
	\bibliographystyle{ieee}
	\bibliography{vgan}
}

\clearpage





\appendix

\input{appendix.tex}

\end{document}

%% file: intro.tex

\begin{abstract}
Visual signals in a video can be divided into content and motion. While content specifies which objects are in the video, motion describes their dynamics. Based on this prior, we propose the Motion and Content decomposed Generative Adversarial Network (MoCoGAN) framework for video generation. The proposed framework generates a video by mapping a sequence of random vectors to a sequence of video frames. Each random vector consists of a content part and a motion part. While the content part is kept fixed, the motion part is realized as a stochastic process. To learn motion and content decomposition in an unsupervised manner, we introduce a novel adversarial learning scheme utilizing both image and video discriminators. Extensive experimental results on several challenging datasets with qualitative and quantitative comparison to the state-of-the-art approaches, verify effectiveness of the proposed framework. In addition, we show that MoCoGAN allows one to generate videos with same content but different motion as well as videos with different content and same motion. 
\vspace{-3.5mm}
\end{abstract}

\section{Introduction}
\label{sec::intro}

\begin{figure}[t]
	\centering
	\begin{overpic}[width=0.47\textwidth]{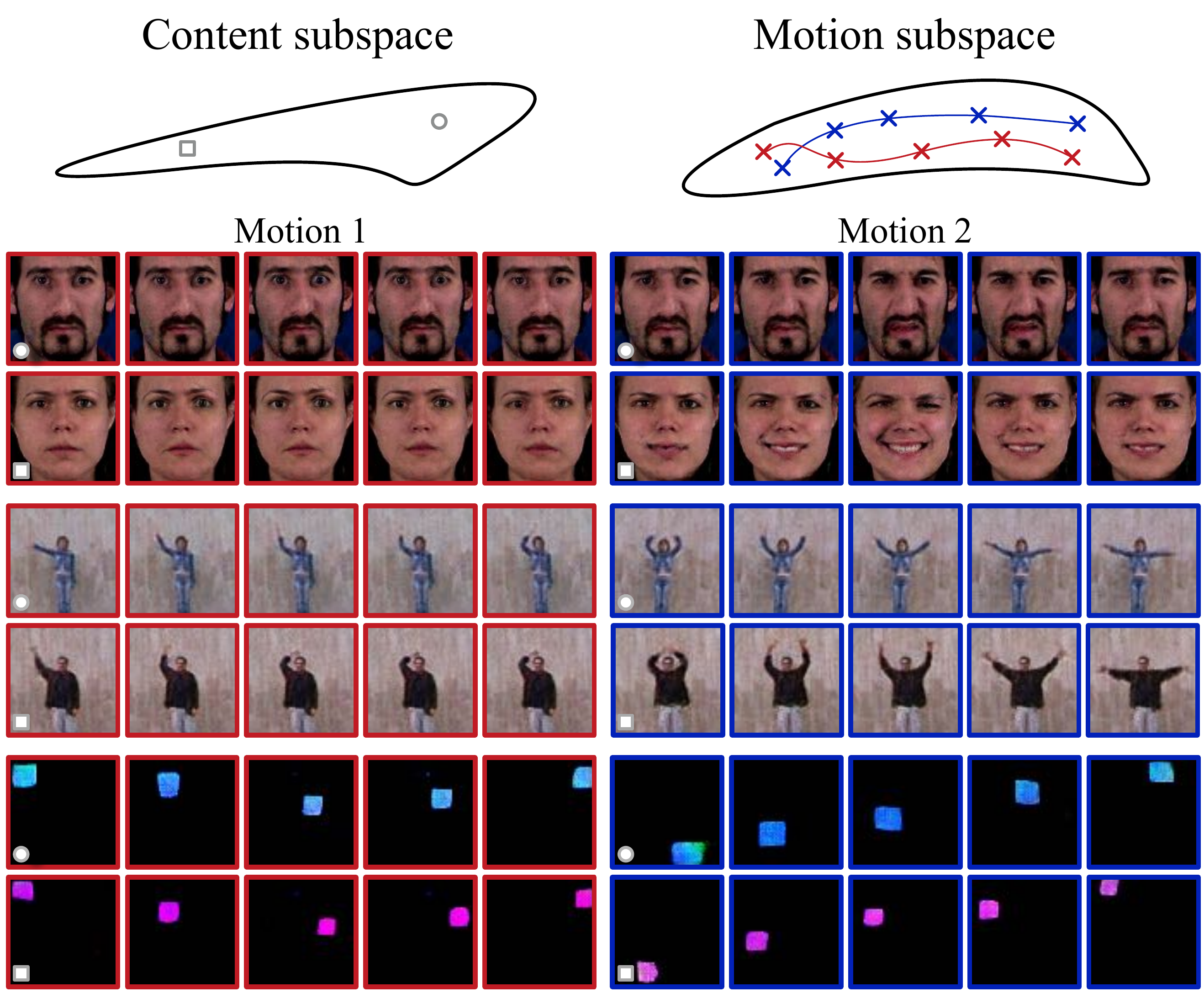}		
	\end{overpic}
	\vspace{-1mm}
	\caption{MoCoGAN adopts a motion and content decomposed representation for video generation. It uses an image latent space (each latent code represents an image) and divides the latent space into content and motion subspaces. By sampling a point in the content subspace and sampling different trajectories in the motion subspace, it generates videos of the same object performing different motion. By sampling different points in the content subspace and the same motion trajectory in the motion subspace, it generates videos of different objects performing the same motion.}
	\label{fig::motion_content_separation}
 	\vspace{-4mm}
\end{figure}

Deep generative models have recently received an increasing amount of attention, not only because they provide a means to learn deep feature representations in an unsupervised manner that can potentially leverage all the unlabeled images on the Internet for training, but also because they can be used to generate novel images necessary for various vision applications. As steady progress toward better image generation is made, it is also important to study the video generation problem. However, the extension from generating images to generating videos turns out to be a highly challenging task, although the generated data has just one more dimension -- the time dimension. 

We argue video generation is much harder for the following reasons. First, since a video is a spatio-temporal recording of visual information of objects performing various actions, a generative model needs to learn the plausible physical motion models of objects in addition to learning their appearance models. If the learned object motion model is incorrect, the generated video may contain objects performing physically impossible motion. Second, the time dimension brings in a huge amount of variations. Consider the amount of speed variations that a person can have when performing a squat movement. Each speed pattern results in a different video, although the appearances of the human in the videos are the same. Third, as human beings have evolved to be sensitive to motion, motion artifacts are particularly perceptible. 


Recently, a few attempts to approach the video generation problem were made through generative adversarial networks (GANs)~\cite{goodfellow2014generative}. Vondrick \etal~\cite{vondrick2016generating} hypothesize that a video clip is a point in a latent space and proposed a VGAN framework for learning a mapping from the latent space to video clips. A similar approach was proposed in the TGAN work~\cite{saito2016temporal}. We argue that assuming a video clip is a point in the latent space unnecessarily increases the complexity of the problem, because videos of the same action with different execution speed are represented by different points in the latent space. Moreover, this assumption forces every generated video clip to have the same length, while the length of real-world video clips varies. An alternative (and likely more intuitive and efficient) approach would assume a latent space of images and consider that a video clip is generated by traversing the points in the latent space. Video clips of different lengths correspond to latent space trajectories of different lengths. 

In addition, as videos are about objects (content) performing actions (motion), the latent space of images should be further decomposed into two subspaces, where the deviation of a point in the first subspace (the content subspace) leads content changes in a video clip and the deviation in the second subspace (the motion subspace) results in temporal motions. Through this modeling, videos of an action with different execution speeds will only result in different traversal speeds of a trajectory in the motion space. Decomposing motion and content allows a more controlled video generation process. By changing the content representation while fixing the motion trajectory, we have videos of different objects performing the same motion. By changing motion trajectories while fixing the content representation, we have videos of the same object performing different motion as illustrated in Fig.~\ref{fig::motion_content_separation}. 

In this paper, we propose the Motion and Content decomposed Generative Adversarial Network \mbox{(MoCoGAN)} framework for video generation. It generates a video clip by sequentially generating video frames. At each time step, an image generative network maps a random vector to an image. The random vector consists of two parts where the first is sampled from a content subspace and the second is sampled from a motion subspace. Since content in a short video clip usually remains the same, we model the content space using a Gaussian distribution and use the same realization to generate each frame in a video clip. On the other hand, sampling from the motion space is achieved through a recurrent neural network where the network parameters are learned during training. Despite lacking supervision regarding the decomposition of motion and content in natural videos, we show that \mbox{MoCoGAN} can learn to disentangle these two factors through a novel adversarial training scheme. Through extensive qualitative and quantitative experimental validations with comparison to the state-of-the-art approaches including VGAN~\cite{vondrick2016generating} and TGAN~\cite{saito2016temporal}, as well as the future frame prediction methods including Conditional-VGAN (C-VGAN)~\cite{vondrick2016generating} and Motion and Content Network (MCNET)~\cite{villegas2017decomposing}, we verify the effectiveness of \mbox{MoCoGAN}.

%% file: rel.tex

\subsection{Related Work}

Video generation is not a new problem. Due to limitations in computation, data, and modeling tools, early video generation works focused on generating dynamic texture patterns~\cite{szummer1996temporal,wei2000fast,doretto2003dynamic}. In the recent years, with the availability of GPUs, Internet videos, and deep neural networks, we are now better positioned to tackle this intriguing problem.

Various deep generative models were recently proposed for image generation including GANs~\cite{goodfellow2014generative}, variational autoencoders (VAEs)~\cite{kingma2013auto,rezende2014stochastic, tulyakov2017hybrid}, and PixelCNNs~\cite{van2016conditional}. In this paper, we propose the MoCoGAN framework for video generation, which is based on GANs. 

Multiple GAN-based image generation frameworks were proposed. Denton~\etal~\cite{denton2015deep} showed a Laplacian pyramid implementation. Radford~\etal~\cite{radford2015unsupervised} used a deeper convolution network. Zhang~\etal~\cite{zhang2016stackgan} stacked two generative networks to progressively render realistic images. Coupled GANs~\cite{liu2016coupled} learned to generate corresponding images in different domains, later extended to translate an image from one domain to a different domain in an unsupervised fashion~\cite{liu2017unsupervised}. InfoGAN~\cite{chen2016infogan} learned a more interpretable latent representation. Salimans~\etal\cite{salimans2016improved} proposed several GAN training tricks. The WGAN~\cite{arjovsky2017wasserstein} and LSGAN~\cite{mao2017least} frameworks adopted alternative distribution distance metrics for more stable adversarial training. Roth~\etal~\cite{roth2017stabilizing} proposed a special gradient penalty to further stabilize training. Karras~\etal~\cite{karras2017progressive} used progressive growing of the discriminator and the generator to generate high resolution images. The proposed MoCoGAN framework generates a video clip by sequentially generating images using an image generator. The framework can easily leverage advances in image generation in the GAN framework for improving the quality of the generated videos. As discussed in Section~\ref{sec::intro},~\cite{vondrick2016generating,saito2016temporal} extended the GAN framework to the video generation problem by assuming a latent space of video clips where all the clips have the same length.

Recurrent neural networks for image generation were previously explored in~\cite{gregor2015draw,im2016generating}. Specifically, some works used recurrent mechanisms to iteratively refine a generated image. Our work is different to~\cite{gregor2015draw,im2016generating} in that we use the recurrent mechanism to generate motion embeddings of video frames in a video clip. The image generation is achieved through a convolutional neural network.

The future frame prediction problem studied in~\cite{srivastava2015unsupervised,oh2015action,mathieu2015deep,kalchbrenner2016video,finn2016unsupervised,van2017transformation,xue2016probabilistic,villegas2017decomposing, denton2017unsupervised} is different to the video generation problem. In future frame prediction, the goal is to predict future frames in a video given the observed frames in the video. Previous works on future frame prediction can be roughly divided into two categories where one focuses on generating raw pixel values in future frames based on the observed ones~\cite{srivastava2015unsupervised,oh2015action,mathieu2015deep,kalchbrenner2016video,xue2016probabilistic,villegas2017decomposing}, while the other focuses on generating transformations for reshuffling the pixels in the previous frames to construct future frames~\cite{finn2016unsupervised,van2017transformation}. The availability of previous frames makes future frame prediction a conditional image generation problem, which is different to the video generation problem where the input to the generative network is only a vector drawn from a latent space. We note that~\cite{villegas2017decomposing} used a convolutional LSTM~\cite{hochreiter1997long} encoder to encode temporal differences between consecutive previous frames for extracting motion information and a convolutional encoder to extract content information from the current image. The concatenation of the motion and content information was then fed to a decoder to predict future frames.

\subsection{Contributions}

\noindent Our contributions are as follows:
\begin{enumerate}[leftmargin=*,itemsep=2pt,topsep=4pt]
    \item We propose a novel GAN framework for unconditional video generation, mapping noise vectors to videos.
    \item We show the proposed framework provides a means to control content and motion in video generation, which is absent in the existing video generation frameworks.
    \item We conduct extensive experimental validation on benchmark datasets with both quantitative and subjective comparison to the state-of-the-art video generation algorithms including VGAN\cite{vondrick2016generating} and TGAN~\cite{saito2016temporal} to verify the effectiveness of the proposed algorithm.
\end{enumerate}

%% file: body.tex
\section{Generative Adversarial Networks}\label{sec::gan}

GANs~\cite{goodfellow2014generative} consist of a generator and a discriminator. The objective of the generator is to generate images resembling real images, while the objective of the discriminator is to distinguish real images from generated ones.

Let $\mathbf{x}$ be a real image drawn from an image distribution, $p_{X}$, and $\mathbf{z}$ be a random vector in $Z_{\mathrm{I}}\equiv\mathbb{R}^d$. Let $G_{\mathrm{I}}$ and $D_{\mathrm{I}}$ be the image generator and the image discriminator. The generator takes $\mathbf{z}$ as input and outputs an image, $\tilde{\mathbf{x}}=G_{\mathrm{I}}(\mathbf{z})$, that has the same support as $\mathbf{x}$. We denote the distribution of $G_{\mathrm{I}}(\mathbf{z})$ as $p_{G_{\mathrm{I}}}$. The discriminator estimates the probability that an input image is drawn from $p_{X}$. Ideally, $D_{\mathrm{I}}(\mathbf{x})=1$ if $\mathbf{x}\sim p_{X}$ and $D_{\mathrm{I}}(\tilde{\mathbf{x}})=0$ if $\tilde{\mathbf{x}}\sim p_{G_{\mathrm{I}}}$. Training of $G_{\mathrm{I}}$ and $D_{\mathrm{I}}$ is achieved via solving a minimax problem given by 
\begin{eqnarray}
\max_{G_{\mathrm{I}}} \min_{D_{\mathrm{I}}} \mathcal{F}_{\mathrm{I}}(D_{\mathrm{I}},G_{\mathrm{I}})
\label{eqn::gan_problem}
\end{eqnarray}
where the functional $\mathcal{F}_{\mathrm{I}}$ is given by 
\begin{eqnarray}
\mathcal{F}_{\mathrm{I}}(D_{\mathrm{I}},G_{\mathrm{I}})
\!\!\!& = &\!\!\! \mathbb{E}_{ \mathbf{x} \sim p_{X}} [ - \log D_{\mathrm{I}}(\mathbf{x}) ] + \nonumber\\
\!\!\!&   &\!\!\!\mathbb{E}_{\mathbf{z}\sim p_{Z_{\mathrm{I}}}} [ - \log( 1 - D_{\mathrm{I}}(G_{\mathrm{I}}(\mathbf{z}))) ].
\end{eqnarray}
In practice,~(\ref{eqn::gan_problem})~is solved by alternating gradient update.

Goodfellow \etal~\cite{goodfellow2014generative} show that, given enough capacity to $D_{\mathrm{I}}$ and $G_{\mathrm{I}}$ and sufficient training iterations, the distribution $p_{G_{\mathrm{I}}}$ converges to $p_{X}$. As a result, from a random vector input $\mathbf{z}$, the network $G_{\mathrm{I}}$ can synthesize an image that resembles one drawn from the true distribution, $p_X$.

\subsection{Extension to Fixed-length Video Generation}

Recently, \cite{vondrick2016generating} extended the GAN framework to video generation by proposing a Video GAN (VGAN) framework. Let $\mathbf{v}^{\mathrm{L}}=[\mathbf{x}^{(1)},...,\mathbf{x}^{(\mathrm{L})}]$ be a video clip with $\mathrm{L}$ frames. The video generation in VGAN is achieved by replacing the vanilla CNN-based image generator and discriminator, $G_{\mathrm{I}}$ and $D_{\mathrm{I}}$, with a spatio-temporal CNN-based video generator and discriminator, $G_{\mathrm{V^L}}$ and $D_{\mathrm{V^L}}$. The video generator $G_{\mathrm{V^L}}$ maps a random vector $\mathbf{z}\in Z_{\mathrm{V^L}}\equiv\mathbb{R}^d$ to a fixed-length video clip, $\tilde{\mathbf{v}}^\mathrm{L}=[\tilde{\mathbf{x}}^{(1)},...,\tilde{\mathbf{x}}^{(\mathrm{L})}]=G_{\mathrm{V^L}}(\mathbf{z})$  and the video discriminator $D_{\mathrm{V^L}}$ differentiates real video clips from generated ones. Ideally, $D_{\mathrm{V^L}}(\mathbf{v}^\mathrm{L})=1$ if $\mathbf{v}^\mathrm{L}$ is sampled from $p_{V^\mathrm{L}}$ and $D_{\mathrm{V^L}}(\tilde{\mathbf{v}}^\mathrm{L})=0$ if $\tilde{\mathbf{v}}^\mathrm{L}$ is sampled from the video generator distribution $p_{G_{\mathrm{V^L}}}$. The TGAN framework~\cite{saito2016temporal} also maps a random vector to a fixed length clip. The difference is that TGAN maps the random vector, representing a fixed-length video, to a fixed number of random vectors, representing individual frames in the video clip and uses an image generator for generation. Instead of using the vanilla GAN framework for minimizing the Jensen-Shannon divergence, the TGAN training is based on the WGAN framework~\cite{arjovsky2017wasserstein} and minimizes the earth mover distance.

\section{Motion and Content Decomposed GAN}\label{sec::MoCoGAN}

In MoCoGAN, we assume a latent space of images $Z_{\mathrm{I}}\equiv\mathbb{R}^d$ where each point $\mathbf{z}\in Z_{\mathrm{I}}$ represents an image, and a video of $\mathrm{K}$ frames is represented by a path of length $\mathrm{K}$ in the latent space, $[\mathbf{z}^{(1)},...,\mathbf{z}^{(\mathrm{K})}]$. By adopting this formulation, videos of different lengths can be generated by paths of different lengths. Moreover, videos of the same action executed with different speeds can be generated by traversing the same path in the latent space  with different speeds.

We further assume $Z_{\mathrm{I}}$ is decomposed into the content $Z_\mathrm{C}$ and motion $Z_\mathrm{M}$ subspaces: $Z_{\mathrm{I}}=Z_\mathrm{C}\times Z_\mathrm{M}$ where $Z_\mathrm{C} = \mathbb{R}^{d_\mathrm{C}}$, $Z_\mathrm{M} = \mathbb{R}^{d_\mathrm{M}}$, and $d=d_\mathrm{C}+d_\mathrm{M}$. The content subspace models motion-independent appearance in videos, while the motion subspace models motion-dependent appearance in videos. For example, in a video of a person smiling, content represents the identity of the person, while motion represents the changes of facial muscle configurations of the person. A pair of the person's identity and the facial muscle configuration represents a face image of the person. A sequence of these pairs represents a video clip of the person smiling. By swapping the look of the person with the look of another person, a video of a different person smiling is represented.

We model the content subspace using a Gaussian distribution: $\mathbf{z}_{\mathrm{C}}\sim p_{Z_\mathrm{C}}\equiv\mathcal{N}(\mathbf{z}|0,I_{d_{\mathrm{C}}})$ where $I_{d_{\mathrm{C}}}$ is an identity matrix of size $d_{\mathrm{C}}\times d_{\mathrm{C}}$. Based on the observation that the content remains largely the same in a short video clip, we use the same realization, $\mathbf{z}_{\mathrm{C}}$, for generating different frames in a video clip. Motion in the video clip is modeled by a path in the motion subspace $Z_\mathrm{M}$. The sequence of vectors for generating a video is represented by 
\begin{equation}
[\mathbf{z}^{(1)},...,\mathbf{z}^{(K)}] = 
\Big{[}
\big{[}
\begin{array}{l}
	\mathbf{z}_\mathrm{C}\\
	\mathbf{z}_{\mathrm{M}}^{(1)}
\end{array}
\big{]},
...,
\big{[}
\begin{array}{l}
	\mathbf{z}_\mathrm{C}\\
	\mathbf{z}_{\mathrm{M}}^{(K)}
\end{array}
\big{]}
\Big{]}
\end{equation}
where $\mathbf{z}_\mathrm{C}\in Z_\mathrm{C}$ and $\mathbf{z}_{\mathrm{M}}^{(k)}\in Z_\mathrm{M}$ for all $k$'s. Since not all paths in $Z_\mathrm{M}$ correspond to physically plausible motion, we need to learn to generate valid paths. We model the path generation process using a recurrent neural network. 

Let $R_{\mathrm{M}}$ to be a recurrent neural network. At each time step, it takes a vector sampled from a Gaussian distribution as input: $\mathbf{\epsilon}^{(k)}\sim p_{\mathrm{E}}\equiv\mathcal{N}(\mathbf{\epsilon}|0,I_{d_\mathrm{E}})$ and outputs a vector in $Z_\mathrm{M}$, which is used as the motion representation. Let $R_{\mathrm{M}}(k)$ be the output of the recurrent neural network at time $k$. Then, $\mathbf{z}_{\mathrm{M}}^{(k)}=R_{\mathrm{M}}(k)$. Intuitively, the function of the recurrent neural network is to map a sequence of independent and identically distributed (i.i.d.) random variables $[\mathbf{\epsilon}^{(1)},...,\mathbf{\epsilon}^{(K)}]$ to a sequence of correlated random variables $[R_{\mathrm{M}}(1),...,R_{\mathrm{M}}(K)]$ representing the dynamics in a video. Injecting noise at every iteration models uncertainty of the future motion at each timestep. We implement $R_{\mathrm{M}}$ using a one-layer GRU network~\cite{chung2014empirical}. 

\begin{figure}[t]
    \centering
	\begin{overpic}[width=0.43\textwidth]{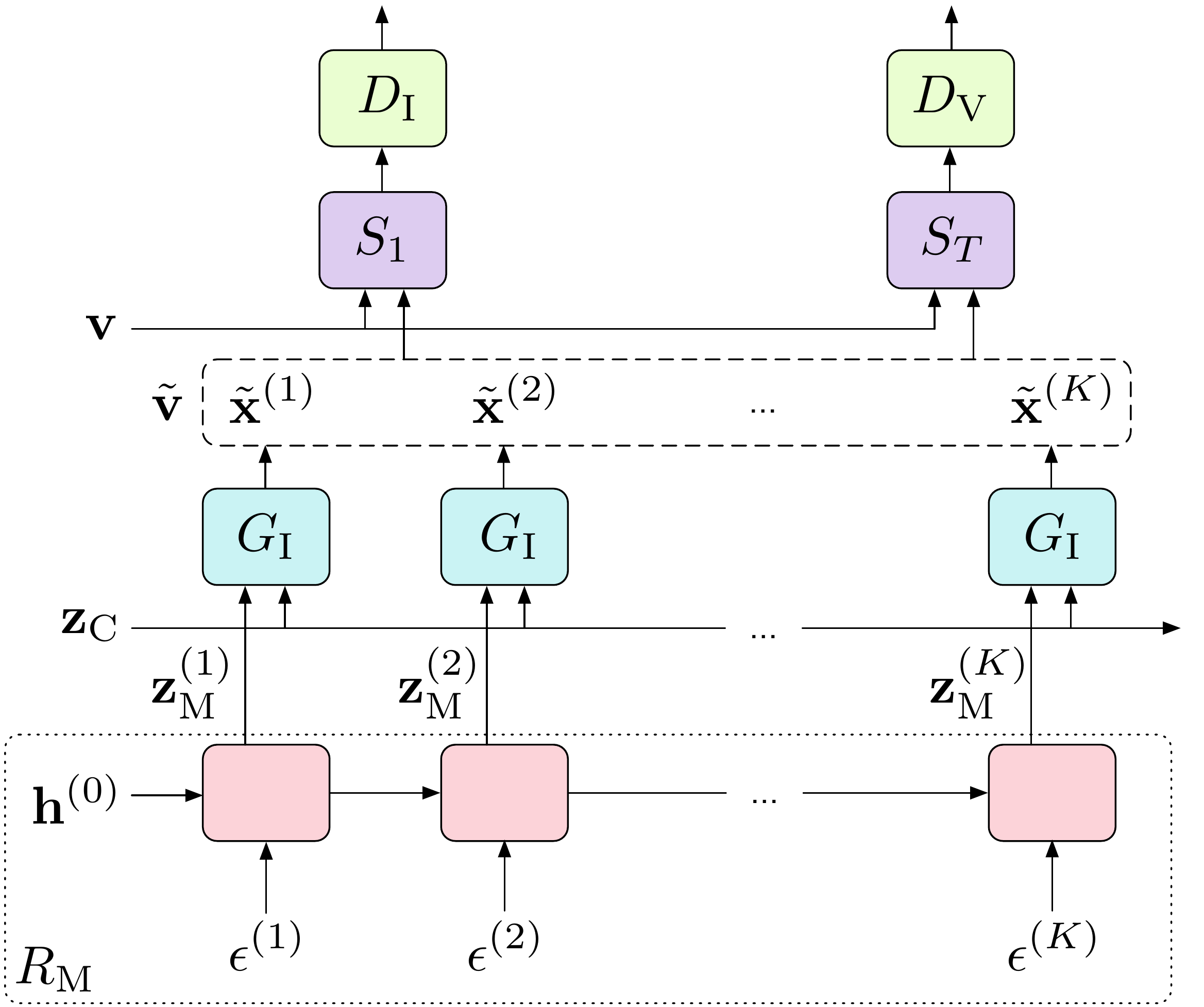}
	\end{overpic}
	\vspace{-2mm}
    	\caption{The \ours framework for video generation. For a video, the content vector, $\mathbf{z}_\mathrm{C}$, is sampled once and fixed. Then, a series of random variables $[\epsilon^{(1)}, ..., \epsilon^{(K)}]$ is sampled and mapped to a series of motion codes $[\mathbf{z}_{\mathrm{M}}^{(1)}, ..., \mathbf{z}_{\mathrm{M}}^{(K)}]$ via the recurrent neural network $R_{\mathrm{M}}$. A generator $G_\mathrm{I}$ produces a frame, $\tilde{\mathbf{x}}^{(k)}$, using the content and the motion vectors $\{\mathbf{z}_{\mathrm{C}}, \mathbf{z}_{\mathrm{M}}^{(k)}\}$. The discriminators, $D_{\mathrm{I}}$ and $D_{\mathrm{V}}$, are trained on real and fake images and videos, respectively, sampled from the training set $\mathbf{v}$ and the generated set $\tilde{\mathbf{v}}$. The function $S_1$ samples a single frame from a video, $S_\mathrm{T}$ samples $\mathrm{T}$ consequtive frames.
	}
	\label{fig::MoCoGAN_network}
	\vspace{-3mm}
\end{figure}

\paragraph{Networks.} MoCoGAN consists of 4 sub-networks, which are the recurrent neural network, $R_{\mathrm{M}}$, the image generator, $G_{\mathrm{I}}$, the image discriminator, $D_{\mathrm{I}}$, and the video discriminator, $D_{\mathrm{V}}$. The image generator generates a video clip by sequentially mapping vectors in $Z_{\mathrm{I}}$ to images, from a sequence of vectors $[\bfrac{\mathbf{z}_{\mathrm{C}}}{\mathbf{z}_{\mathrm{M}}^{(1)}},...,\bfrac{\mathbf{z}_{\mathrm{C}}}{\mathbf{z}_{\mathrm{M}}^{(K)}}]$ to a sequence of images, $\tilde{\mathbf{v}}=[\tilde{\mathbf{x}}^{(1)},...,\tilde{\mathbf{x}}^{(K)}]$, where $\tilde{\mathbf{x}}^{(k)}=G_{\mathrm{I}}(\bfrac{\mathbf{z}_{\mathrm{C}}}{\mathbf{z}_{\mathrm{M}}^{(k)}})$ and $\mathbf{z}_{\mathrm{M}}^{(k)}$'s are from the recurrent neural network, $R_{\mathrm{M}}$. We note that the video length $K$ can vary for each video generation.

Both $D_{\mathrm{I}}$ and $D_{\mathrm{V}}$ play the judge role, providing criticisms to $G_{\mathrm{I}}$ and $R_{\mathrm{M}}$. The image discriminator $D_{\mathrm{I}}$ is specialized in criticizing $G_{\mathrm{I}}$ based on individual images. It is trained to determine if a frame is sampled from a real video clip, $\mathbf{v}$, or from $\tilde{\mathbf{v}}$. On the other hand, $D_{\mathrm{V}}$ provides criticisms to $G_{\mathrm{I}}$ based on the generated video clip. $D_{\mathrm{V}}$ takes a fixed length video clip, say $\mathrm{T}$ frames, and decides if a video clip was sampled from a real video or from $\tilde{\mathrm{v}}$. Different from $D_{\mathrm{I}}$, which is based on vanilla CNN architecture, $D_{\mathrm{V}}$ is based on a spatio-temporal CNN architecture. We note that the clip length $\mathrm{T}$ is a hyperparameter, which is set to 16 throughout our experiments. We also note that $\mathrm{T}$ can be smaller than the generated video length $K$. A video of length $K$ can be divided into $K-\mathrm{T}+1$ clips in a sliding-window fashion, and each of the clips can be fed into $D_{\mathrm{V}}$.

The video discriminator $D_{\mathrm{V}}$ also evaluates the generated motion. Since $G_{\mathrm{I}}$ has no concept of motion, the criticisms on the motion part go directly to the recurrent neural network, $R_{\mathrm{M}}$. In order to generate a video with realistic dynamics that fools $D_{\mathrm{V}}$, $R_{\mathrm{M}}$ has to learn to generate a sequence of motion codes $[\mathbf{z}_\mathrm{M}^{(1)},...,\mathbf{z}_\mathrm{M}^{(K)}]$ from a sequence of i.i.d.\ noise inputs $[\epsilon^{(1)},...,\epsilon^{(K)}]$ in a way such that $G_{\mathrm{I}}$ can map $\mathbf{z}^{(k)}=[{\mathbf{z}_\mathrm{C}},\medspace {\mathbf{z}_\mathrm{M}^{(k)}}]$
to consecutive frames in a video.

Ideally, $D_{\mathrm{V}}$ alone should be sufficient for training $G_{\mathrm{I}}$ and $R_\mathrm{M}$, because $D_{\mathrm{V}}$ provides feedback on both static image appearance and video dynamics. However, we found that using $D_{\mathrm{I}}$ significantly improves the convergence of the adversarial training. This is because  training $D_{\mathrm{I}}$ is simpler, as it only needs to focus on static appearances. Fig.~\ref{fig::MoCoGAN_network} shows visual representation of the MoCoGAN framework.

\paragraph{Learning.} Let $p_{\mathrm{V}}$ be the distribution of video clips of variable lengths. Let $\kappa$ be a discrete random variable denoting the length of a video clip sampled from $p_{\mathrm{V}}$. (In practice, we can estimate the distribution of $\kappa$, termed $p_{\mathrm{K}}$, by computing a histogram of video clip length from training data). To generate a video, we first sample a content vector, $\mathbf{z}_{\mathrm{C}}$, and a length, $\kappa$. We then run $R_{\mathrm{M}}$ for $\kappa$ steps and, at each time step, $R_{\mathrm{M}}$ takes a random variable $\mathbf{\epsilon}$ as the input. A generated video is then given by
\begin{equation}
\tilde{\mathbf{v}}=\Bigg{[} G_{\mathrm{I}}(\bfrac{z_{\mathrm{C}}}{R_{\mathrm{M}}(1)}), ...,G_{\mathrm{I}}(\bfrac{z_{\mathrm{C}}}{R_{\mathrm{M}}(\kappa)})\Bigg{]}.
\end{equation}

Recall that our $D_{\mathrm{I}}$ and $D_{\mathrm{V}}$ take one frame and $\mathrm{T}$ consecutive frames in a video as input, respectively. In order to represent these sampling mechanisms, we introduce two random access functions $S_1$ and $S_\mathrm{T}$. The function $S_1$ takes a video clip (either $\mathbf{v}\sim p_{\mathrm{V}}$ or $\tilde{\mathbf{v}}\sim p_{\tilde{\mathrm{V}}}$) and outputs a random frame from the clip, while the function $S_\mathrm{T}$ takes a video clip and randomly returns $\mathrm{T}$ consecutive frames from the clip. With this notation, the \mbox{MoCoGAN} learning problem is
\vspace{-2mm}
\begin{align}
\max_{G_{\mathrm{I}},R_{\mathrm{M}}} \min_{D_{\mathrm{I}},D_{\mathrm{V}}} \mathcal{F}_{\mathrm{V}}(D_{\mathrm{I}},D_{\mathrm{V}},G_{\mathrm{I}},R_{\mathrm{M}})
\label{eqn::mocogan_problem}
\end{align}
where the objective function $\mathcal{F}_{\mathrm{V}} (D_{\mathrm{I}}, D_{\mathrm{V}},G_{\mathrm{I}},R_{\mathrm{M}})$ is 
\begin{eqnarray}
\mathbb{E}_{\mathbf{v}} [ - \log D_{\mathrm{I}}(S_1(\mathbf{v})) ] +
                    \mathbb{E}_{\tilde{\mathbf{v}}} [-\log(1-D_{\mathrm{I}}(S_1 (\tilde{\mathbf{v}})))] + \nonumber\\
\mathbb{E}_{\mathbf{v}} [ - \log D_{\mathrm{V}}(S_\mathrm{T}(\mathbf{v})) ] + 
                    \mathbb{E}_{\tilde{\mathbf{v}}} [-\log(1-D_{\mathrm{V}}(S_\mathrm{T} (\tilde{\mathbf{v}})))],
\label{eqn::f_v}
\end{eqnarray}
where $\mathbb{E}_{\mathbf{v}}$ is a shorthand for $\mathbb{E}_{\mathbf{v} \sim p_{\mathrm{V}}}$, and $\mathbb{E}_{\tilde{\mathbf{v}}}$ for $\mathbb{E}_{\tilde{\mathbf{v}} \sim p_{\tilde{\mathrm{V}}}}$. In (\ref{eqn::f_v}), the first and second terms encourage $D_{\mathrm{I}}$ to output 1 for a video frame from a real video clip $\mathbf{v}$ and 0 for a video frame from a generated one $\tilde{\mathbf{v}}$. Similarly, the third and fourth terms encourage $D_{\mathrm{V}}$ to output 1 for $\mathrm{T}$ consecutive frames in a real video clip $\mathbf{v}$ and 0 for $\mathrm{T}$ consecutive frames in a generated one $\tilde{\mathbf{v}}$. 
The second and fourth terms encourage the image generator and the recurrent neural network to produce realistic images and video sequences of $\mathrm{T}$-consecutive frames, such that no discriminator can distinguish them from real images and videos.

We train MoCoGAN using the alternating gradient update algorithm as in~\cite{goodfellow2016nips}. Specifically, in one step, we update $D_{\mathrm{I}}$ and $D_{\mathrm{V}}$ while fixing $G_{\mathrm{I}}$ and $R_{\mathrm{M}}$. In the alternating step, we update $G_{\mathrm{I}}$ and $R_{\mathrm{M}}$ while fixing $D_{\mathrm{I}}$ and $D_{\mathrm{V}}$.

\subsection{Categorical Dynamics}\label{sec::category}

Dynamics in videos are often categorical (e.g., discrete action categories: walking, running, jumping, etc.). To model this categorical signal, we augment the input to $R_{\mathrm{M}}$ with a categorical random variable, $\mathbf{z}_{\mathrm{A}}$, where each realization is a one-hot vector. We keep the realization fixed since the action category in a short video remains the same. The input to $R_{\mathrm{M}}$ is then given by 
\begin{equation}
\Bigg{[}
\bigg{[}
\begin{array}{l}
	\mathbf{z}_\mathrm{A}\\
	\mathbf{\epsilon}^{(1)}
\end{array}
\bigg{]},
...,
\bigg{[}
\begin{array}{l}
	\mathbf{z}_\mathrm{A}\\
	\mathbf{\epsilon}^{(K)}
\end{array}
\bigg{]}
\Bigg{]},
\end{equation}
To relate $\mathbf{z}_{\mathrm{A}}$ to the true action category, we adopt the InfoGAN learning~\cite{chen2016infogan} and augment the objective function in (\ref{eqn::f_v}) to   
$\mathcal{F}_{\mathrm{V}}(D_{\mathrm{I}},D_{\mathrm{V}},G_{\mathrm{I}},R_{\mathrm{M}})+\lambda L_{I}(G_{\mathrm{I}},Q)$
where $L_I$ is a lower bound of the mutual information between the generated video clip and $\mathbf{z}_{\mathrm{A}}$, $\lambda$ is a hyperparameter, and the auxiliary distribution $Q$ (which approximates the distribution of the action category variable conditioning on the video clip) is implemented by adding a softmax layer to the last feature layer of $D_{\mathrm{V}}$. We use $\lambda=1$. We note that when the labeled training data are available, we can train $Q$ to output the category label for a real input video clip to further improve the performance~\cite{45830}. 

%% file: expr_refined.tex
\section{Experiments}

We conducted extensive experimental validation to evaluate MoCoGAN. In addition to comparing to VGAN~\cite{vondrick2016generating} and TGAN~\cite{saito2016temporal}, both quantitatively and qualitatively, we evaluated the ability of MoCoGAN to generate 1) videos of the same object performing different motions by using a fixed content vector and varying motion trajectories and 2) videos of different objects performing the same motion by using different content vectors and the same motion trajectory. We then compared a variant of the MoCoGAN framework with state-of-the-art next frame prediction methods: VGAN and MCNET~\cite{villegas2017decomposing}. Evaluating generative models is known to be a challenging task~\cite{theis2015note}. Hence, we report experimental results on several datasets, where we can obtain reliable performance metrics:


\begin{itemize}[leftmargin=*,itemsep=-1pt,topsep=0pt]
\item \textbf{Shape motion.} The dataset contained two types of shapes (circles and squares) with varying sizes and colors, performing two types of motion: one moving from left to right, and the other moving from top to bottom. The motion trajectories were sampled from Bezier curves. There were $4,000$ videos in the dataset; the image resolution was $64\times64$ and video length was 16. 
\item \textbf{Facial expression.} We used the MUG Facial Expression Database~\cite{aifanti2010mug} for this experiment. The dataset consisted of $86$ subjects. Each video consisted of $50$ to $160$ frames. We cropped the face regions and scaled to $96\times96$. We discarded videos containing fewer than $64$ frames and used only the sequences representing one of the six  facial expressions: anger, fear, disgust, happiness, sadness, and surprise. In total, we trained on $1,254$ videos.
\item \textbf{Tai-Chi.} We downloaded $4,500$ Tai Chi video clips from YouTube. For each clip, we applied a human pose estimator~\cite{cao2017realtime} and cropped the clip so that the performer is in the center. Videos were scaled to $64\times64$ pixels.
\item \textbf{Human actions.} We used the Weizmann Action data\-base~\cite{gorelick2007actions}, containing $81$ videos of 9 people performing 9 actions, including jumping-jack and waving-hands. We scaled the videos to $96\times96$. Due to the small size, we did not conduct a quantitative evaluation using the dataset. Instead, we provide visual results in Fig.~\ref{fig::motion_content_separation} and Fig.~\ref{fig::weizman-action}. 
\item \textbf{UCF101}~\cite{soomro2012ucf101}{\bf.} The database is commonly used for video action recognition. It includes $13,220$ videos of 101 different action categories. Similarly to the TGAN work \cite{saito2016temporal}, we scaled each frame to $85 \times 64$ and cropped the central $64 \times 64$ regions for learning.
\end{itemize}

\paragraph{Implementation.} The details of the network designs are given in the supplementary materials. We used ADAM~\cite{kingma2014adam} for training, with a learning rate of 0.0002 and momentums of 0.5 and 0.999. Our code will be made public.

\subsection{Video Generation Performance}

\begin{figure*}[h]
	\begin{subfigure}[t]{0.33\textwidth}
	    \begin{minipage}{\textwidth}
	            \makebox[0mm][s]{
        			\includegraphics[width=\textwidth]{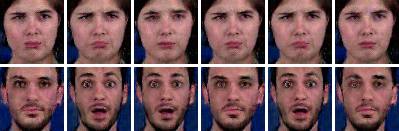}
        		}
        		\animategraphics[width=\textwidth]{10}{figures/new-videos/mocogan/faces/}{00}{47}
		\end{minipage}    	
		\begin{minipage}{\textwidth}
    		\makebox[0mm][s]{
    			\includegraphics[width=\textwidth]{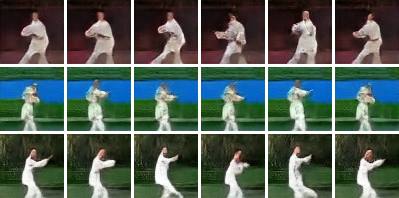}
    		}
    		\animategraphics[width=\textwidth]{10}{figures/new-videos/mocogan/taichi/}{00}{47}
		\end{minipage}
		\caption{Generated by MoCoGAN}
	\end{subfigure}
	~
	\begin{subfigure}[t]{0.33\textwidth}
	    \begin{minipage}{\textwidth}
        		\makebox[0mm][s]{
        			\includegraphics[width=\textwidth]{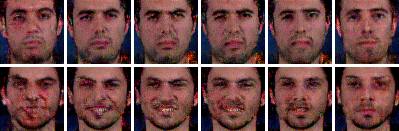}
        		}
        		\animategraphics[width=\textwidth]{10}{figures/new-videos/vgan/faces/}{00}{31}
		\end{minipage}    	
		\begin{minipage}{\textwidth}
    		\makebox[0mm][s]{
    			\includegraphics[width=\textwidth]{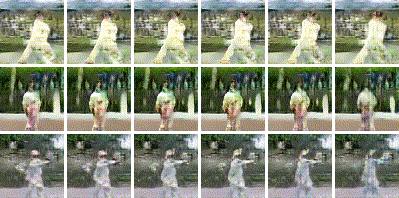}
    		}
    		\animategraphics[width=\textwidth]{5}{figures/new-videos/vgan/taichi/}{00}{16}
		\end{minipage}
		\caption{Generated by VGAN~\cite{vondrick2016generating}}
	\end{subfigure}
	~
	\begin{subfigure}[t]{0.33\textwidth}
	    \begin{minipage}{\textwidth}
        		\makebox[0mm][s]{
        			\includegraphics[width=\textwidth]{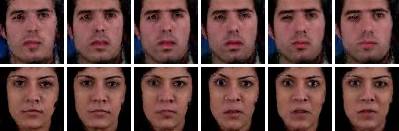}
        		}
        		\animategraphics[width=\textwidth]{10}{figures/new-videos/tgan/faces/}{00}{15}
		\end{minipage}    	
		\begin{minipage}{\textwidth}
    		\makebox[0mm][s]{
    			\includegraphics[width=\textwidth]{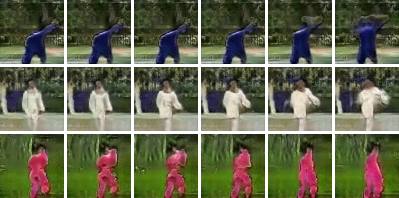}
    		}
    		\animategraphics[width=\textwidth]{5}{figures/new-videos/tgan/taichi/}{00}{15}
		\end{minipage}
		\caption{Generated by TGAN~\cite{saito2016temporal}}
	\end{subfigure}
	\vspace{-2mm}
	\caption{Generated video clips used in the user study. The video clips were randomly selected. The figure is best viewed via the Acrobat Reader on a desktop. Click each image to play the video clip.}
	\label{fig::pairwise_comparisons}
	\vspace{-4mm}
\end{figure*}

\begin{table}[t!]
\centering
\caption{Video generation content consistency comparison. A smaller ACD means the generated frames in a video are perceptually more similar. We also compute the ACD for the training set, which is the reference.}
\label{tab::comparison2vgan}\vspace{-2mm}
\centering
\begin{tabularx}{210pt}{lcc}
\toprule
    ACD                                   & Shape Motion & Facial Expressions \tabularnewline\midrule
    Reference                             & 0            & 0.116              \tabularnewline
    VGAN~\cite{vondrick2016generating}    & 5.02         & 0.322              \tabularnewline
    TGAN~\cite{saito2016temporal}         & 2.08         & {0.305}            \tabularnewline
    MoCoGAN                               & {\bf 1.79}  & {\bf 0.201}         \tabularnewline\bottomrule
\end{tabularx}
\end{table}

\paragraph{Quantitative comparison.} We compared MoCoGAN to VGAN and TGAN\footnote{The VGAN and TGAN implementations are provided by their authors.} using the shape motion and facial expression datasets. For each dataset, we trained a video generation model and generated 256 videos for evaluation. The VGAN and TGAN implementations can only generate fixed-length videos (32 frames and 16 frames correspondingly). For a fair comparison, we generated 16 frames using MoCoGAN, and selected every second frame from the videos generated by VGAN, such that each video has 16 frames in total.

For quantitative comparison, we measured content consistency of a generated video using the Average Content Distance (ACD) metric. For shape motion, we first computed the average color of the generated shape in each frame. Each frame was then represented by a 3-dimensional vector. The ACD is then given by the average pairwise L2 distance of the per-frame average color vectors. For facial expression videos, we employed \mbox{OpenFace}~\cite{amos2016openface}, which outperforms human performance in the face recognition task, for measuring video content consistency. OpenFace produced a feature vector for each frame in a face video. The ACD was then computed using the average pairwise L2 distance of the per-frame feature vectors.

We computed the average ACD scores for the 256 videos generated by the competing algorithms for comparison. The results are given in Table~\ref{tab::comparison2vgan}. From the table, we found that the content of the videos generated by MoCoGAN was more consistent, especially for the facial expression video generation task: MoCoGAN achieved an ACD score of 0.201, which was almost 40\% better than 0.322 of VGAN and 34\% better than 0.305 of TGAN. Fig.~\ref{fig::pairwise_comparisons} shows examples of facial expression videos for competing algorithms.

\begin{table}[t!]
\caption{Inception score for models trained on UCF101. All values except MoCoGAN's are taken from~\cite{saito2016temporal}.}
\label{table-exp-inception}\vspace{-2mm}
\centering
\begin{tabularx}{220pt}{lccc}
	\toprule
	{} 				                    & VGAN & TGAN & MoCoGAN		                    \\
	\midrule
    UCF101                                              & $8.18 \pm .05$ & $11.85 \pm .07$ & $\textbf{12.42} \pm \textbf{.03}$  \\
	\bottomrule
\end{tabularx}
\vspace{-2mm}
\end{table}

\begin{table}[t!]
\caption{User preference score on video generation quality.}
\label{table-exp-user-study}\vspace{-2mm}
\centering
\begin{tabularx}{210pt}{lcc}
	\toprule
	{User preference, \%}             & Facial Exp.          & Tai-Chi              \\
	\midrule
	MoCoGAN / VGAN	   	              & {\bf 84.2} / 15.8    & {\bf 75.4} / 24.6    \\
	MoCoGAN / TGAN                    & {\bf 54.7} / 45.3    & {\bf 68.0} / 32.0    \\
	\bottomrule
\end{tabularx}
\end{table}


Furthermore, we compared with TGAN and VGAN by training on the UCF101 database and computing the inception score similarly to~\cite{saito2016temporal}. Table~\ref{table-exp-inception} shows comparison results. In this experiment we used the same MoCoGAN model as in all other experiments. We observed that MoCoGAN was able to learn temporal dynamics better, due to the decomposed representation, as it generated more realistic temporal sequences. We also noted that TGAN reached the inception score of 11.85 with WGAN training procedure and Singular Value Clipping (SVC), while MoCoGAN showed a higher inception score 12.42 without such tricks, supporting that the proposed framework is more stable than and superior to the TGAN approach.

\paragraph{User study.} We conducted a user study to quantitatively compare MoCoGAN to VGAN and TGAN using the facial expression and Tai-Chi datasets. For each algorithm, we used the trained model to randomly generate 80 videos for each task. We then randomly paired the videos generated by the MoCoGAN with the videos from one of the competing algorithms to form 80 questions. These questions were sent to the workers on Amazon Mechanical Turk (AMT) for evaluation. The videos from different algorithms were shown in random order for a fair comparison. Each question was answered by 3 different workers. The workers were instructed to choose the video that looks more realistic. Only the workers with a lifetime HIT (Human Intelligent Task) approval rate greater than 95\% participated in the user study. 

We report the average preference scores (the average number of times, a worker prefers an algorithm) in Table~\ref{table-exp-user-study}. From the table, we find that the workers considered the videos generated by MoCoGAN more realistic most of the times. Compared to VGAN, MoCoGAN achieved a preference score of 84.2\% and 75.4\% for the facial expression and Tai-Chi datasets, respectively. Compared to TGAN, MoCoGAN achieved a preference score of 54.7\% and 68.0\% for the facial expression and Tai-Chi datasets, respectively. In Fig.~\ref{fig::pairwise_comparisons}, we visualize the facial expression and Tai-Chi videos generated by the competing algorithms. We find that the videos generated by MoCoGAN are more realistic and contained less content and motion artifacts.

\begin{figure}[t!]
    \begin{subfigure}[t]{0.49\textwidth}
        \centering
        \makebox[0mm][s]{
        		\includegraphics[width=0.976\textwidth]{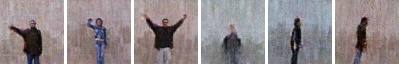}
        	}	
    	\animategraphics[width=0.976\textwidth]{10}{figures/animated/actions2/0018_}{0000}{047}
        \vspace{-1.5mm}     
        \caption{Samples from the model trained on the Weizmann database.}
        \label{fig::weizman-action}
        \vspace{1mm}        
    \end{subfigure}
    
    \vspace{-1mm}

    \begin{subfigure}[t]{0.49\textwidth}
        \centering
    	\makebox[0mm][s]{
    		\includegraphics[width=0.976\textwidth]{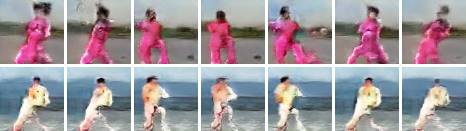}
    	}
    	\animategraphics[width=0.97\textwidth]{10}{figures/animated/content-motion-separation-taichi/}{00}{31}
    	\vspace{-1.5mm} 
        \caption{Examples of changing the motion code while fixing the content code. Every row has fixed content, every column has fixed motion.}
        \label{fig::content-motion-separation}
        \vspace{1mm} 
    \end{subfigure}
    
    \vspace{-1mm}
    
    \begin{subfigure}[t]{0.49\textwidth}
        \centering
    	\makebox[0mm][s]{
    		\includegraphics[width=0.97\textwidth]{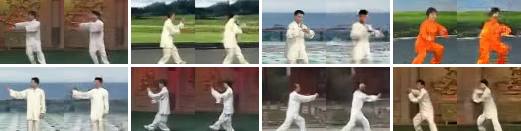}
    	}
    	\animategraphics[width=0.97\textwidth]{10}{figures/animated/conditional-taichi/}{00}{15}
    	\vspace{-1.5mm} 
        \caption{Image-to-video translation. In each block: input image (left), video generated by MoCoGAN (right).}\label{fig::image-to-video-translation}
        \vspace{1mm} 
    \end{subfigure}
    \vspace{-4mm}
    \caption{The figure is best viewed with Acrobat Reader on a desktop. Click each image to play the video clip.}
    \vspace{-3mm}
\end{figure}

\paragraph{Qualitative evaluation.} We conducted a qualitative experiment to demonstrate our motion and content decomposed representation. We sampled two content codes and seven motion codes, giving us 14 videos. Fig.~\ref{fig::content-motion-separation} shows an example randomly selected from this experiment. Each row has the same content code, and each column has the same motion code. We observed that MoCoGAN generated the same motion sequences for two different content samples.

\subsection{Categorical Video Generation}

We augmented MoCoGAN with categorical variables and trained it for facial expression video generation as described in Section~\ref{sec::category}. The MUG dataset contains 6 different facial expressions and hence $\mathbf{z}_{\mathrm{A}}$ is realized as a 6 dimensional one-hot vector. We then generated 96 frames of facial expression videos. During generation, we changed the action category, $\mathbf{z}_\mathrm{A}$, every 16 frames to cover all 6 expressions. Hence, a generated video corresponded to a person performing 6 different facial expressions, one after another.

To evaluate the performance, we computed the ACD of the generated videos. A smaller ACD means the generated faces over the 96 frames were more likely to be from the same person. Note that the ACD reported in this subsection are generally larger than the ACD reported in Table~\ref{tab::comparison2vgan}, because the generated videos in this experiment are 6 times longer and contain 6 facial expressions versus 1. We also used the motion control score (MCS) to evaluate MoCoGAN's capability in motion generation control. To compute MCS, we first trained a spatio-temporal CNN classifier for action recognition using the labeled training dataset. 
During test time, we used the classifier to verify whether the generated video contained the action. The MCS is then given by testing accuracy of the classifier. A model with larger MCS offers better control over the action category.

In this experiment, we also evaluated the impact of different conditioning schemes to the categorical video generation performance. The first scheme is our default scheme where $\mathbf{z}_\mathrm{A}\rightarrow R_\mathrm{M}$. The second scheme, termed $\mathbf{z}_\mathrm{A}\rightarrow G_\mathrm{I}$, was to feed the category variable directly to the image generator. In addition, to show the impact of the image discriminative network $D_\mathrm{I}$, we considered training the MoCoGAN framework without $D_\mathrm{I}$. 

\begin{figure*}[t!]
		\centering
		\includegraphics[width=0.95\textwidth]{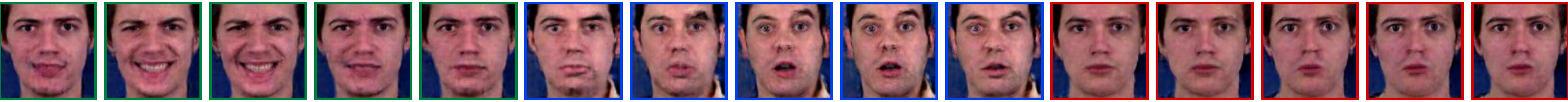}
		\includegraphics[width=0.95\textwidth]{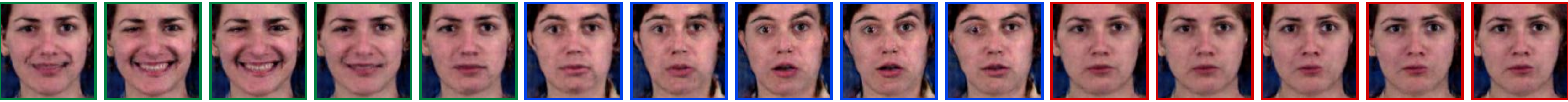}

    	\caption{Generated videos of changing facial expressions. We changed the expression from smile to fear through surprise.}
    	\label{fig::expression_control}
 	    \vspace{-4mm}
\end{figure*}

\begin{table}[t!]
\centering
\caption{Performance on categorical facial expression video generation with various MoCoGAN settings.}
\label{table-exp-1-2}
\begin{tabular}{lllcc}
	\toprule
	\multicolumn{2}{c}{Settings} & MCS & ACD \\
	\midrule
	 \sout{$D_I$}	& $\mathbf{z}_\mathrm{A} \rightarrow G_\mathrm{I}$ & 0.472 & 1.115	 \\
	 \sout{$D_I$}	& $\mathbf{z}_\mathrm{A} \rightarrow R_\mathrm{M}$ & 0.491 & 1.073	  \\
	 $D_I$	        & $\mathbf{z}_\mathrm{A} \rightarrow G_\mathrm{I}$ & 0.355 & 0.738  \\
	 $D_I$	        & $\mathbf{z}_\mathrm{A} \rightarrow R_\mathrm{M}$ & {\bf 0.581 } & {\bf 0.606}  \\
	\bottomrule
\end{tabular}
\end{table}

Table~\ref{table-exp-1-2} shows experimental results. We find that the models trained with $D_\mathrm{I}$ consistently yield better performances on various metrics. We also find that $\mathbf{z}_\mathrm{A}\rightarrow R_\mathrm{M}$ yields better performance. Fig.~\ref{fig::expression_control} shows two videos from the best model in Table~\ref{table-exp-1-2}. We observe that by fixing the content vector but changing the expression label, it generates videos of the same person performing different expressions. And similarly, by changing the content vector and providing the same motion trajectory, we generate videos of different people showing the same expression sequence.

\begin{figure}
	\centering
	\includegraphics[width=0.48\textwidth]{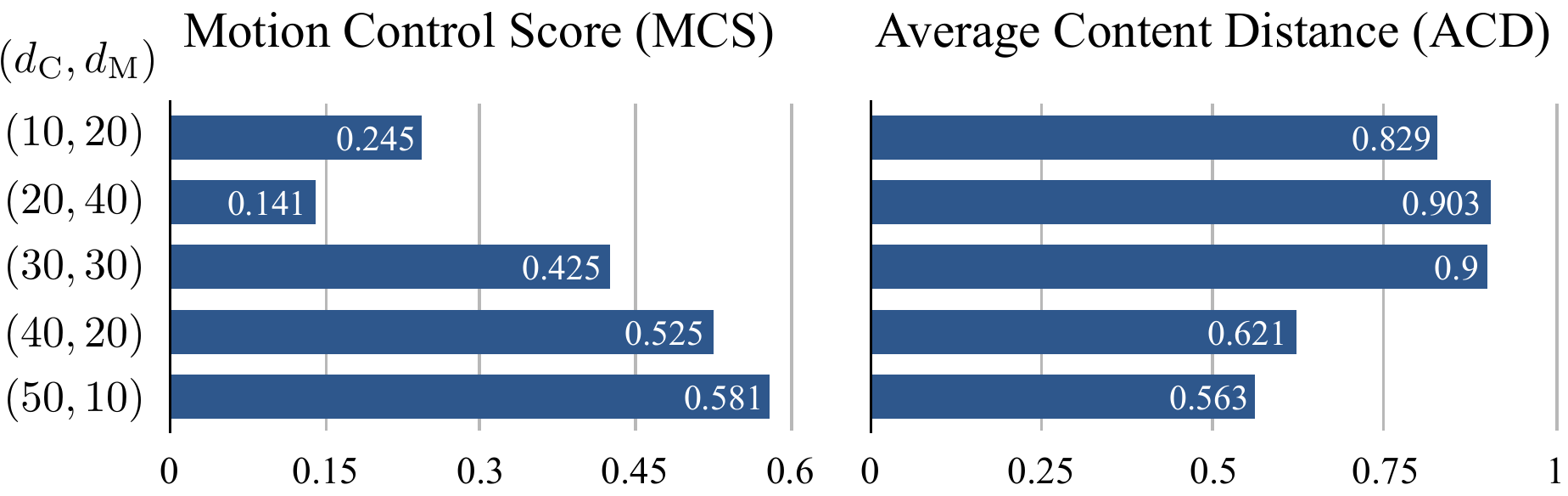}
	\vspace{-2mm}
	\caption{MoCoGAN models with varying $(d_{\mathrm{C}},d_{\mathrm{M}})$ settings on facial expression generation.}
	\label{ablation-study}
\end{figure}

We conducted an experiment to empirically analyze the impact of the dimensions of the content and motion vectors $\mathbf{z}_{\mathrm{C}}$ and $\mathbf{z}_\mathrm{M}^{(t)}$ (referred to as $d_{\mathrm{C}}$ and $d_{\mathrm{M}}$) to the categorical video generation performance. In the experiment, we fixed the sum of the dimensions to 60 (i.e., $d_{\mathrm{C}} + d_{\mathrm{M}}=60$) and changed the value of $d_{\mathrm{C}}$ from 10 to 50, with a step size of 10. Fig.~\ref{ablation-study} shows the results. 

We found when $d_\mathrm{C}$ was large, MoCoGAN had a small ACD. This meant a video generated by the MoCoGAN resembled the same person performing different expressions. We were expecting a larger $\mathbf{z}_\mathrm{M}$ would lead to a larger MCS but found the contrary. Inspecting the generated videos, we found when $d_\mathrm{M}$ was large (i.e. $d_\mathrm{C}$ was small), MoCoGAN failed to generate recognizable faces, resulting in a poor MCS. In this case, given poor image quality, the facial expression recognition network could only perform a random guess on the expression and scored poorly. Based on this, we set $d_\mathrm{C}=50$ and $d_\mathrm{M}=10$ in all the experiments.


\subsection{Image-to-video Translation}

\begin{table}[t!]
\caption{User preference score on the quality of the image-to-video-translation results.}
\label{table-exp-video-prediction}\vspace{-2mm}
\centering
\begin{tabularx}{170pt}{lc}
	\toprule
	{User preference, \%}                 & Tai-Chi            \\
	\midrule
	MoCoGAN / C-VGAN	   	              & {\bf 66.9} / 33.1    \\
	MoCoGAN / MCNET	   	                  & {\bf 65.6} / 34.4    \\
	\bottomrule
\end{tabularx}
\end{table}

We trained a variant of the MoCoGAN framework, in which the generator is realized as an encoder-decoder architecture~\cite{liu2017unsupervised}, where the encoder produced the content code $\mathbf{z}_\mathrm{C}$ and the initial motion code $\mathbf{z}_\mathrm{m}^{(0)}$. Subsequent motion codes were produced by $R_\mathrm{M}$ and concatenated with the content code to generate each frame. That is the input was an image and the output was a video. We trained a MoCoGAN model using the Tai-Chi dataset. In test time, we sampled random images from a withheld test set to generate video sequences. In addition to the MoCoGAN loss, we have added the $L_{1}$ reconstruction loss for training the encoder-decoder architecture similar to~\cite{liu2017unsupervised}. Under this setting, MoCoGAN generated a video sequence starting from the first frame (see Fig.~\ref{fig::image-to-video-translation}). We compared with two state-of-the-art approaches on this task: a Conditional-VGAN (C-VGAN) and Motion Content Network (MCNET)~\cite{villegas2017decomposing} and performed a user study to compare the competing methods. We note that MCNET used 4 frames to predict a video, while C-VGAN and MoCoGAN required a single frame only. Table~\ref{table-exp-video-prediction} shows the user preference scores. The results support that MoCoGAN was able to generate more realistic videos than C-VGAN and MCNET. 



%% file: conc.tex

\section{Conclusion}

We presented the MoCoGAN framework for motion and content decomposed video generation. Given sufficient video training data, MoCoGAN automatically learns to disentangle motion from content in an unsupervised manner. For instance, given videos of people performing different facial expressions, MoCoGAN learns to separate a person's identity from their expression, thus allowing us to synthesize a new video of a person performing different expressions, or fixing the expression and generating various identities. 
This is enabled by a new generative adversarial network, which generates a video clip by sequentially generating video frames. Each video frame is generated from a random vector, which consists of two parts, one signifying content and one signifying motion. The content subspace is modeled with a Gaussian distribution, whereas the motion subspace is modeled with a recurrent neural network. We sample this space in order to synthesize each video frame.
Our experimental evaluation supports that the proposed framework is superior to current state-of-the-art video generation and next frame prediction methods.

%% file: appendix.tex
\section{Network Architecture}\label{sec::network_arch}
\vspace{-6pt}
\begin{table}[h]
	\centering
	\caption{Network architectures of the image generative network $G_{\mathrm{I}}$, the image discriminative network $D_{\mathrm{I}}$, and the video generative network $D_{\mathrm{V}}$ used in the experiments.}
	\resizebox{0.46\textwidth}{!}{
		
		\begin{tabular}{cl}
			\toprule[1.5pt]
			$G_{\mathrm{I}}$ & Configuration                                                                                                   \\
			\midrule[1pt]
			Input            & $[ \mathbf{z}_{\mathrm{a}}\sim\mathcal{N}(\mathbf{0}, \mathbf{I}), \mathbf{z}_{\mathrm{m}}\sim R_{\mathrm{M}}]$ \\
			\midrule[0.01pt] 
			0                & DCONV-(N512, K6, S0, P0), BN, LeakyReLU                        \\
			\midrule[0.01pt]
			1                & DCONV-(N256, K4, S2, P1), BN, LeakyReLU                        \\
			\midrule[0.01pt]
			2                & DCONV-(N128, K4, S2, P1), BN, LeakyReLU                        \\
			\midrule[0.01pt]
			3                & DCONV-(N64, K4, S2, P1), BN, LeakyReLU                         \\
			\midrule[0.01pt]
			4                & DCONV-(N3, K4, S2, P1), BN, LeakyReLU                          \\
			\bottomrule[1.5pt]
			{} & {} \\
			
		    \toprule[1.5pt]
			$D_{\mathrm{I}}$ & Configuration                                                  \\
			\midrule[1pt]
			Input            & height $\times$ width $\times$ 3                               \\
			\midrule[0.01pt]
			0                & CONV-(N64, K4, S2, P1), BN, LeakyReLU                          \\
			\midrule[0.01pt]
			1                & CONV-(N128, K4, S2, P1), BN, LeakyReLU                         \\
			\midrule[0.01pt]
			2                & CONV-(N256, K4, S2, P1), BN, LeakyReLU                         \\
			\midrule[0.01pt]
			3                & CONV-(N1, K4, S2, P1), Sigmoid                                 \\
			\bottomrule[1.5pt]
			{} & {} \\
			\toprule[1.5pt]
			$D_{\mathrm{V}}$ & Configuration                                                 \\
			\midrule[1pt]
			Input            & 16 $\times$ height $\times$ width $\times$ 3                  \\
			\midrule[0.01pt]
			0                & CONV3D-(N64, K4, S1, P0), BN, LeakyReLU                       \\
			\midrule[0.01pt]
			1                & CONV3D-(N128, K4, S1, P0), BN, LeakyReLU                      \\
			\midrule[0.01pt]
			2                & CONV3D-(N256, K4, S1, P0), BN, LeakyReLU                      \\
			\midrule[0.01pt]
			3                & CONV3D-(N1, K4, S1, P0), Sigmoid                              \\
			\bottomrule[1.5pt]
		\end{tabular}	
	}
	\label{supplement-configuration-table}
\end{table}
\vspace{-5pt}
Table~\ref{supplement-configuration-table} detailed the network architecture used in the main paper. We used several different type of layers. A convolutional layer with $N$ output channels, kernel size $K$, stride $S$, and padding $P$ is denoted in the table as CONV-($N$, $K$, $S$, $P$) and similarly for 3D convolutional layers CONV3D-($N$, $K$, $S$, $P$). Kernel size, padding, and stride are equal for all the dimensions in each layer. Batch normalization layers are followed by the LeakyReLU nonlinearity in our case. The $R_{\mathrm{M}}$ consisted of a single GRU module.

\vspace{-2pt}
\section{Additional Qualitative Results}
\begin{figure*}[t!]
	\centering
	\includegraphics[width=0.98\textwidth]{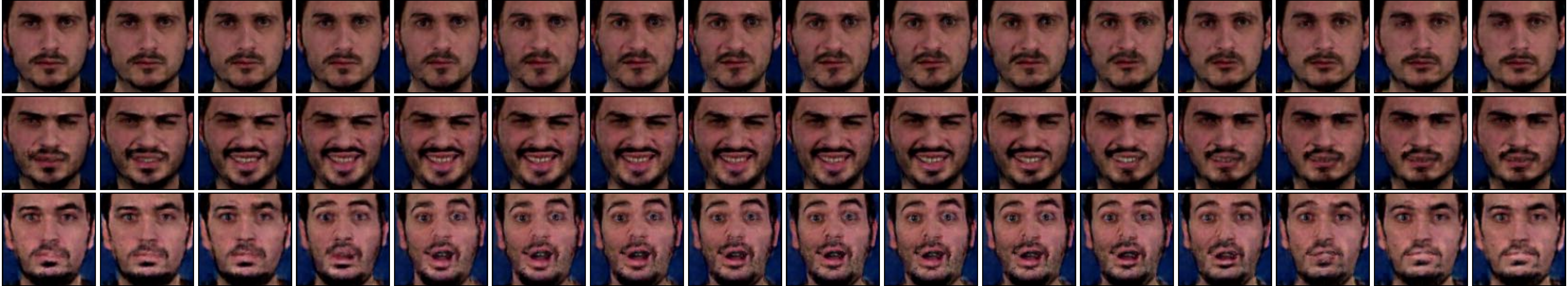}
	
	\vspace{0.5em}
	
	\includegraphics[width=0.98\textwidth]{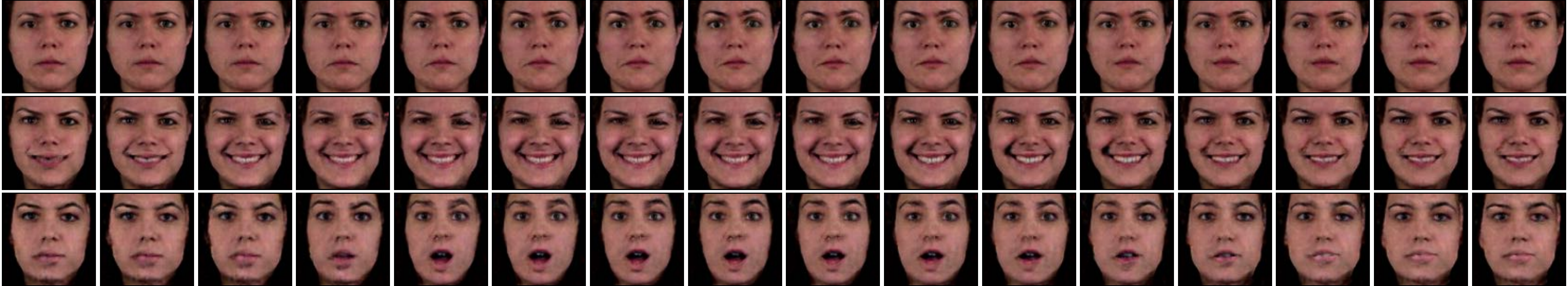}
	
	\vspace{0.5em}
	
	\includegraphics[width=0.98\textwidth]{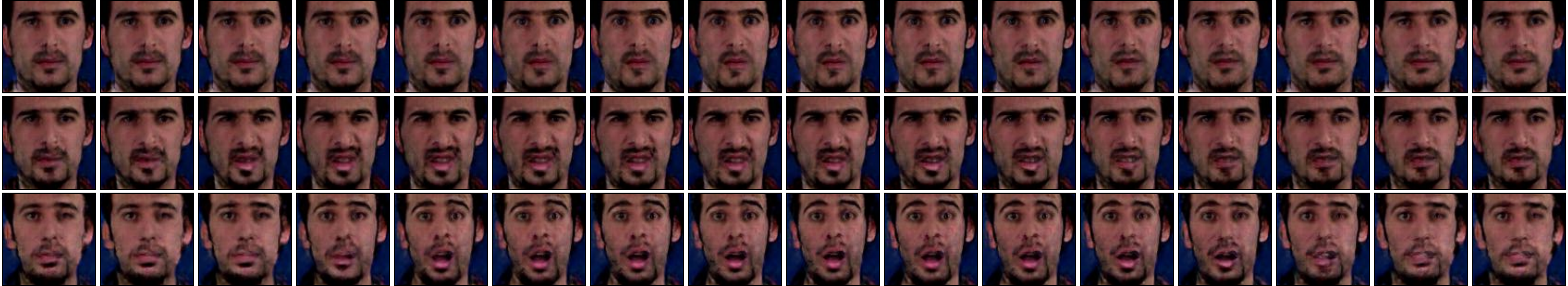}
	
	\vspace{0.5em}
	
	\includegraphics[width=0.98\textwidth]{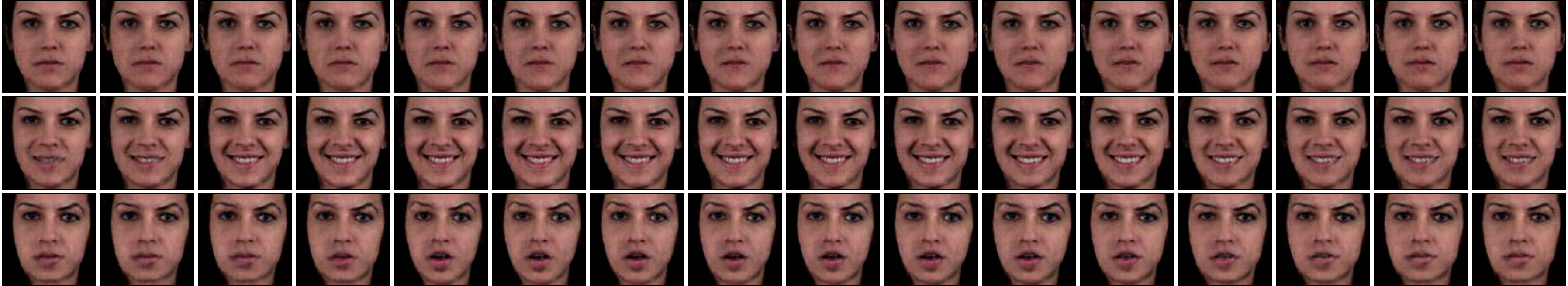}
	
	\vspace{0.5em}
	
	\includegraphics[width=0.98\textwidth]{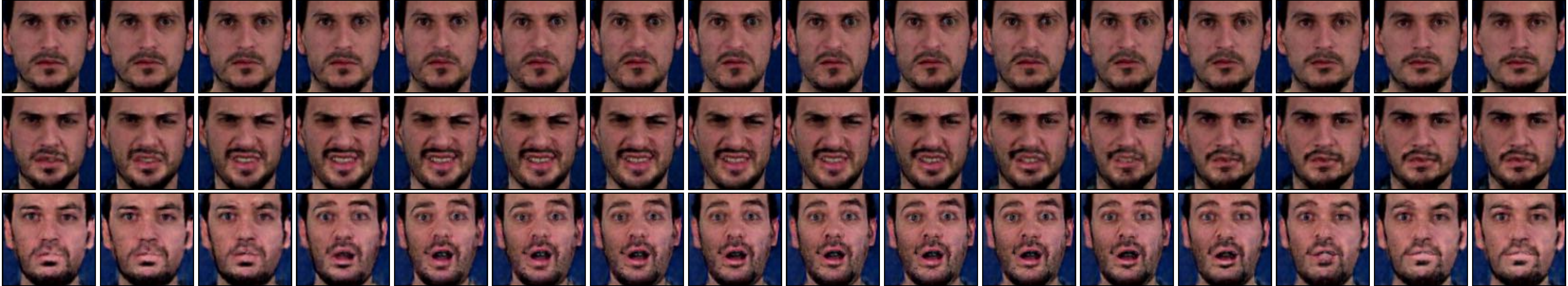}
	
	\caption{Facial expression video generation results. Every three rows have the same identity but different $\mathbf{z}_{\mathrm{A}}$.}
	\label{supplement-static-faces}
\end{figure*}

\begin{figure*}[t!]
	\centering
	
	\includegraphics[width=0.98\textwidth]{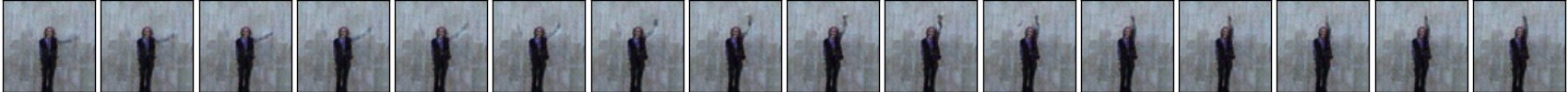}
	
	\includegraphics[width=0.98\textwidth]{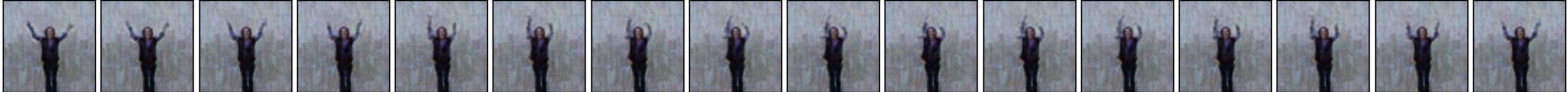}
	
	\includegraphics[width=0.98\textwidth]{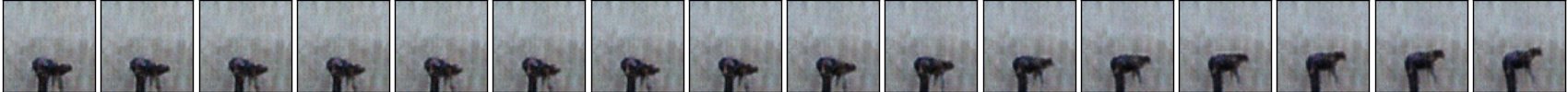}
	
	\vspace{0.5em}
	
	\includegraphics[width=0.98\textwidth]{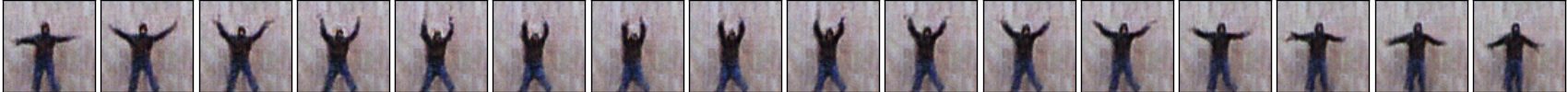}
	
	\includegraphics[width=0.98\textwidth]{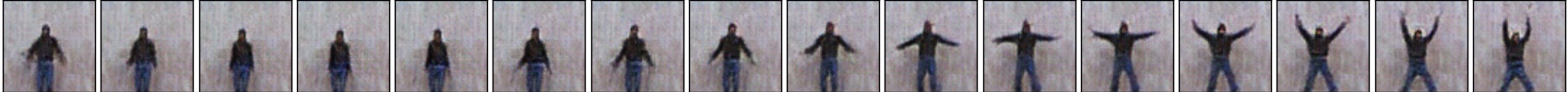}
	
	\includegraphics[width=0.98\textwidth]{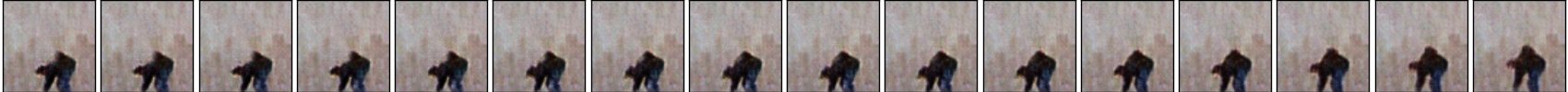}
	
	\vspace{0.5em}
	
	\includegraphics[width=0.98\textwidth]{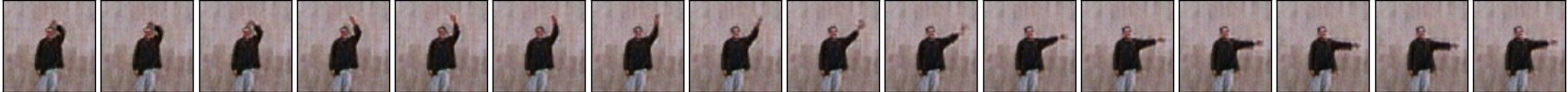}
	
	\includegraphics[width=0.98\textwidth]{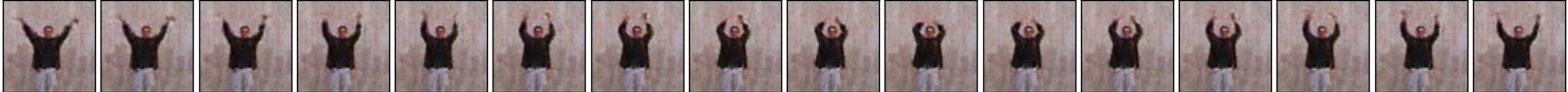}
	
	\includegraphics[width=0.98\textwidth]{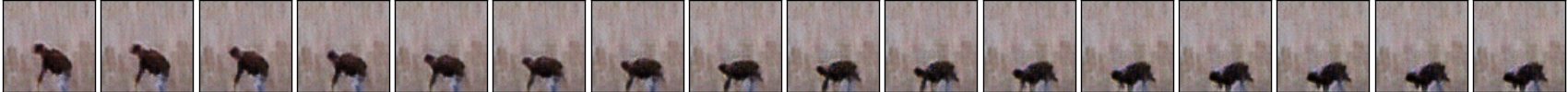}
	
	\vspace{0.5em}
	
	\includegraphics[width=0.98\textwidth]{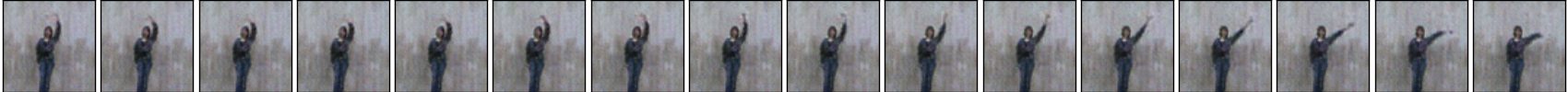}
	
	\includegraphics[width=0.98\textwidth]{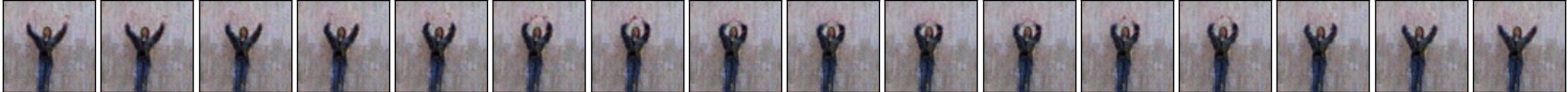}
	
	\includegraphics[width=0.98\textwidth]{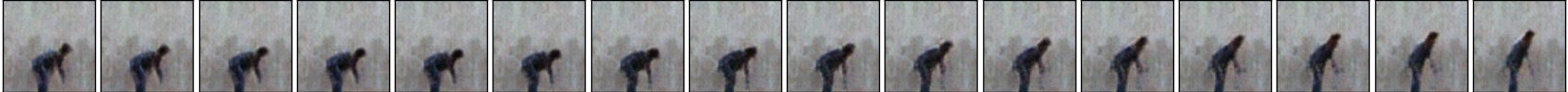}
	
	\vspace{0.5em}
	
	\includegraphics[width=0.98\textwidth]{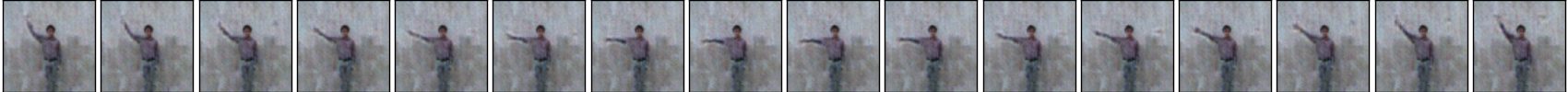}
	
	\includegraphics[width=0.98\textwidth]{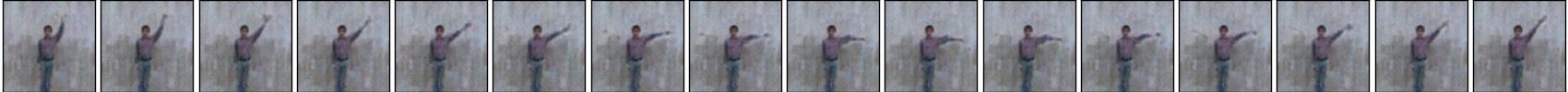}
	
	\includegraphics[width=0.98\textwidth]{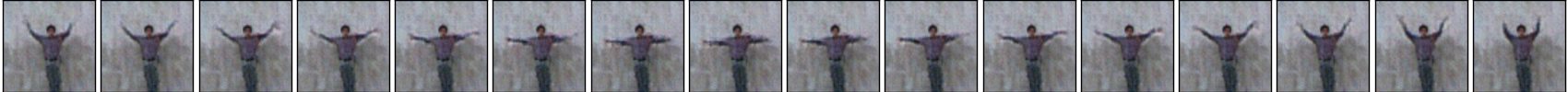}
	
	\caption{Human action video generation results. Every three rows have the same identity but different $\mathbf{z}_{\mathrm{A}}$.}
	\label{supplement-static-actions}
\end{figure*}

\vspace{-2pt}
Figures~\ref{supplement-static-faces} and \ref{supplement-static-actions} show categorical facial expression and categorical human actions video generation results, respectively. In each figure, every group of three rows was generated with a fixed content vector $\mathbf{z}_\mathrm{C}$, but random action vector $\mathbf{z}_\mathrm{A}$ and motion vectors $\mathbf{z}_\mathrm{M}^{(t)}$'s. We found that the identity was kept fixed throughout a video. We noted that the length of the generated videos were longer than those in the training data. This showed that the MoCoGAN can generalize the video generation along the time dimension.

